# Robot can reduce superior's dominance in group discussions with human social hierarchy

Kazuki Komura
SOKENDAI(The Graduate University for Advanced Studies)
Japan
National Institute of Informatics
Japan
kazukikomura@nii.ac.jp

Kumi Ozaki
Department of radiology, Hamamatsu University School of Medicine
Japan
ozaki@hama-med.ac.jp

Seiji Yamada
National Institute of Informatics
Japan
Department of Informatics, SOKENDAI (The Graduate University for Advanced Studies)
Japan
seiji@nii.ac.jp

## Abstract
This study investigated whether robotic agents that deal with social hierarchical relationships can reduce the dominance of superiors and equalize participation among participants in discussions with hierarchical structures. Thirty doctors and students having hierarchical relationship were gathered as participants, and an intervention experiment was conducted using a robot that can encourage participants to speak depending on social hierarchy. These were compared with strategies that intervened equally for all participants without considering hierarchy and with a no-action. The robots performed follow actions, showing backchanneling to speech, and encourage actions, prompting speech from members with less speaking time, on the basis of the hierarchical relationships among group members to equalize participation. The experimental results revealed that the robot's actions could potentially influence the speaking time among members, but it could not be conclusively stated that there were significant differences between the robot's action conditions. However, the results suggested that it might be possible to influence speaking time without decreasing the satisfaction of superiors. This indicates that in discussion scenarios where experienced superiors are likely to dominate, controlling the robot's backchanneling behavior could potentially suppress dominance and equalize participation among group members.

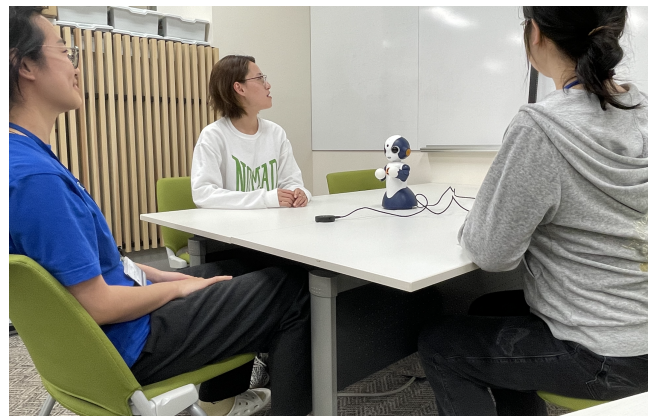

**Figure 1: Screenshot of three participants gathered around a robot conducting a discussion experiment**

## 1 Introduction

In modern society, causal discussion like brainstorming in small groups consisting of two or more individuals is playing an increasingly important role [28].There are various scenarios where small groups engage in problem-solving, such as family discussions, company meetings, and study sessions in educational settings.

Research has been advancing on how problem-solving in these groups is conducted and what factors influence their performance [1]. Many of these studies focus on aspects such as the number of group members, the formation of collective intelligence within the group, confidence, satisfaction, and consensus in decision-making, all of which impact group problem-solving performance and discussion performance [4, 6, 16]. Among these, the dynamic elements within the group have been highlighted as significant predictors of group problem-solving performance. Woolley et al.'s [30] research suggests that the way people interact, particularly the balance of turn-taking in conversations, is a crucial dynamic determinant of overall team performance. In recent years, research on the use of robots and AI to support discussions in small groups has been gaining attention. One focus of these technologies is to facilitate communication within the group and create an environment where all members can actively participate [25, 27, 29].However, most of these studies do not take into account the social relationships and social roles of the members. When humans engage in discussions, they often have social relationships with each other. Just as the content of a conversation can change depending on whom a speaker is addressing, the dynamics of a group can significantly shift depending on these social relationships [12]. Despite this, there are still few studies that have investigated interventions considering the social relationships of the group members.

In particular, hierarchical relationships within a group (e.g.,leader-member, teacher-student) are considered to have a significant impact on group dynamics. In such hierarchical groups, the opinions of superiors tend to be strongly reflected, and these members often dominate the conversation, while subordinates tend to refrain from speaking and are more likely to yield their turns to superiors.When subordinates hesitate to speak up to superiors, their opportunities and time to speak decrease, which can lead to a decline in the overall group dynamics [11, 12]. This tendency is known to be cross-cultural [5]. If anthropomorphic agents can intervene in a manner appropriate to these hierarchical relationships, they could have a significant impact.

In this study, we examined whether the linguistic and non-linguistic behaviours of the conversational robot SOTA could generate discussion dynamics within hierarchical groups. The robot is designed to suppress the dominance of superiors and enhance the equality of participation within the group. This experiment targeted hierarchical groups, considering that strong hierarchical relationships centered around doctors are observed in medical settings and may negatively impact problem-solving [18]. The experiment involved a total of 30 participants, including experienced doctors (superiors) and students or less experienced doctors (subordinates) recruited from a university hospital and a university. Specifically, we developed strategies for the robot that understood the hierarchical structure of the group and examined the impact on the speaking time/number of turns/satisfaction of group members.

## 2 Related Work
### 2.1 Group conversational dynamics
Hackman and Morris [15] assert that an important predictor of problem-solving in group discussions lies in the continuous interaction processes that occur among group members while they are engaged in the task. This indicates that not only the sum of the abilities of each team member but also how and who interacts within the group is crucial for problem-solving. Specifically, Woolley [30] and colleagues argue that it is more important how members exchange knowledge and the frequency of these exchanges rather than the abilities of the members themselves. They predict that performance improves when conversational turn-taking/speaking time within the group is more evenly distributed. The idea that balanced engagement during problem-solving is important is reflected in several other studies, which show that more comprehensive member participation, active discussions, and constructive dissent lead to improved group performance and more innovative outcomes [21]. Researchers have shown that constructive dissent and innovative group performance can be achieved by enabling groups to distribute speaking time/turn-taking more evenly [9].

### 2.2 Group conversational dynamics in hierarchical relationships
It is known that conversations are influenced by the internal structure of the members participating in them. In particular, hierarchical relationships are considered to be a major factor affecting turn-taking and speaking time in conversations [5, 17]. Gibson et al. [12] investigated how hierarchical relationships influence turn-taking and shifts in speaking targets on the basis of actual meeting scenarios. Their findings revealed that superiors tend to assert or seize the right to speak while others are talking. Additionally, subordinates tend to return the right to speak or avoid speaking after a superior has spoken. These results can be interpreted to mean that in groups with social hierarchies, superiors are in an environment where it is easier for them to speak, whereas subordinates are in an environment where they tend to avoid speaking or find it difficult to speak.

Furthermore, Bisel et al. [2] argue that there is a "hierarchical mum effect" between superiors and subordinates, where inconvenient or undesirable information is withheld. This suggests that hierarchical relationships are a factor influencing the ease of speaking, particularly leading to fewer utterances and turns by subordinates. In this case, considering the equity of participation within the group, it is necessary to create situations where subordinates find it easier to speak and to avoid situations where superiors dominate the conversation for extended periods.

### 2.3 Robot that intervene in groups
On the basis of the findings that group dynamics are a crucial factor in the performance and satisfaction of group discussions, several studies in Human-Agent Interaction (HAI) and Human-Robot Interaction (HRI) have investigated whether robots can influence team dynamics. In a study by Tennent et al. [26], it was demonstrated that a microphone-shaped robot could improve group discussion performance by engaging in backchanneling and facilitation behaviours to encourage participation from less active members in a three-person discussion. Correia et al. [7]'s research showed that a robot expressing group-based emotions could enhance trust within the team. Furthermore, a study by Sebo et al. [24] indicated that an anthropomorphic robot making vulnerable statements could elicit more empathetic behaviours from its three teammates engaged in a cooperative problem-solving game. These studies demonstrate that the behaviour of robots within a team can influence group dynamics, and the current research builds upon these insights.

### 2.4 Robots that intervene in groups considering social relationships
The aforementioned studies on robot interventions did not consider the relationships among group members when conducting their interventions. However, it is conceivable that group dynamics can significantly change depending on the nature of social relationships within the group. For instance, if one member is well-versed in a particular topic while the others are not, the knowledgeable member is likely to dominate the conversation. There are also some studies that have considered these social relationships. In the study by Gillet et al. [13, 14], the robot's gaze was examined in relation to conversational imbalances among participants with different languages skills (native speakers and second-language speakers). It was suggested that individuals with higher language skills tend to lead the conversation more, and the robot's gaze influenced interactions among participants. Zhang et al. [31]'s research focused on the individuals meeting for the first time individuals and demonstrated that a robot could intervene in ice-breaking activities, leading to more effective development of close relationships. Neto et al.'s [22] study showed that by intervening more strongly in the communication between visually impaired and non-visually impaired children, it was possible to achieve more equal participation.

Although these studies do not specifically address hierarchical relationships, they highlight the growing interest in experiments targeting social relationships.

## 3 Hypothesis

To our understanding, while there is research on robots/AI intervening in small groups, there is no research yet on agents that understand hierarchical structures and suppress the dominance of superiors. Considering real-world applications, it is ideal for interventions not only to suppress dominance but also to avoid lowering the satisfaction of all participants, as in the study by Kim et al. [19]. On the basis of the above background, we formulated two hypotheses for this study:

**H1:** When a robot takes actions with social hierarchy, the group members' participation level(speaking time/turn) becomes more uniform.

**H2:** When a robot takes actions with social hierarchy, the satisfaction of group members does not decrease.

To answer these questions, our study examines the effects under three conditions: a robot that takes actions considering the social hierarchical structure, a robot that takes actions targeting all members, and a condition where the robot does not move. We will evaluate the following dependent valuables:

(1) Speaking time of superiors or subordinates
(2) Number of turns taken by superiors or subordinates
(3) Overall speaking equity within the group
(4) Overall turn-taking equity within the group
(5) Satisfaction with the discussion of superiors or subordinates

Additionally, task performance is known to be related to task satisfaction [8]. Therefore, we also measured the discussion satisfaction of each member and compared it with the above indicators. Examining the relationship between these and the objective indicators of group discussions is a secondary objective of this study.

## 4 Designing Robot Interaction

### 4.1 Experiment environment

For our research, we used the SOTA(produced by Vstone Co. Ltd.) robot. SOTA is 28 cm tall, has a child-like form, and is equipped with LEDs on its face for low-level emotional expressions and conversation support. SOTA can perform simple gestures with two degrees of freedom in each hand, and its body can also rotate. We developed a system that allows SOTA to take action according to the discussion context. This system consists of two main components: input processing from a microphone array and action instructions to SOTA. The microphone array is placed at the center of the table where the discussion takes place and can record the direction from which sounds are coming. These inputs are processed in real-time through a local PC, which calculates the direction of the speaker. Using speech recognition technology, it helps identify which participant is speaking. SOTA's actions are configured to be executed under specific conditions using HTTP communication. This setup allows SOTA to perform actions in real-time, synchronized with the ongoing discussion, creating an interactive environment.

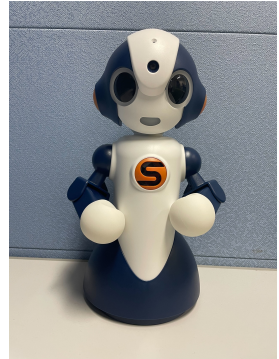

Figure 2: SOTA is 28 cm tall, has a child-like form, and is equipped with conversation support.

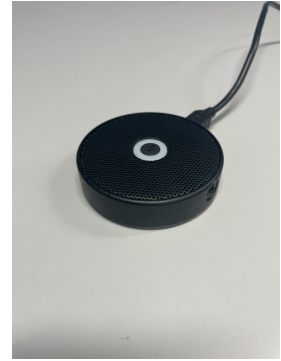

Figure 3: Voice Compass is a small microphone module with 12 directional microphones.

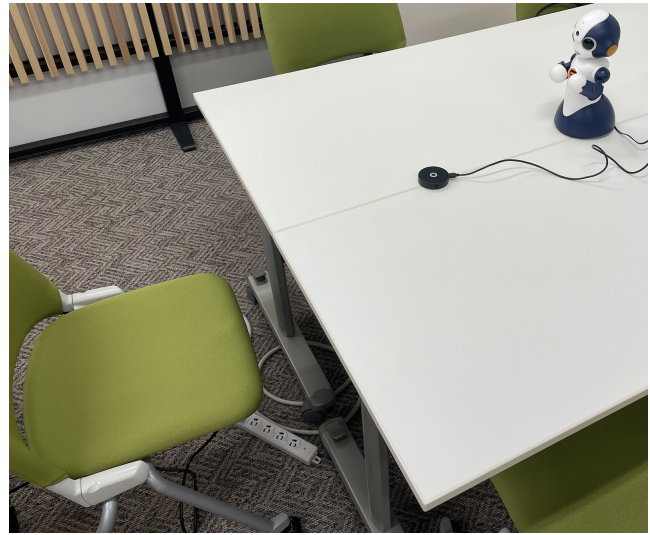

Figure 4: Laboratory environment. In front of robot, a microphone array is placed to capture direction from which participants are speaking.

### 4.2 Robot behaviour

The robot was designed to have two main actions. The first is a follow action that responds to the person speaking, which is a form of backchanneling. The second is an encourage action that prompts the participant who has spoken the least to speak. Both actions are designed to promote participation in the discussion, with the aim of actively continuing the conversation (follow) and encouraging speech (encourage).

In the follow action, the robot rotated towards the person speaking. The input from the microphone array detected speech in 0.1-second intervals, and to prevent reacting to backchanneling or sudden noises, the robot was set to act only when speech from the same direction was detected in $n$ out of $m$ consecutive 0.1-second intervals. Through pilot studies conducted in the laboratory, $n$ and $m$ were empirically determined to be 3 and 2, respectively. Additionally, on the basis of findings from previous research, the action was designed to occur with a 150-ms delay after detecting speech to avoid any sense of discomfort.

In the encourage action, the robot prompted the participant who had spoken the least to speak more every minute. The robot rotated towards the participant who had spoken the least, opened both hands to encourage speech, and said, "What do you think?" To determine which of the 10 actions was most effective in promoting speech, an online pilot study was conducted in advance, and the action with the highest speech promotion score was adopted.

## 5 Evaluation of Robot Behaviour

### 5.1 Study design

We designed a within-subjects study with three conditions to evaluate the effects of a robot that understands hierarchical structures, on the dynamics and satisfaction of group interpersonal relationships. An a priori power analysis using G*Power [10] indicated that at least eight groups were needed for a high effect size ($f$ = .40) with the power at .80 and alpha at .05. A plan was made to gather a total of 10 groups and 30 participants from these results.

Participants were recruited from medical institutions where hierarchical relationships in the workplace are clear and can potentially cause negative issues in problem-solving [18]. To create a social hierarchy $\boxed{\text{Doctor}} \xleftarrow{up} \boxed{\text{Student}}$ within the groups, participants were recruited from a group with little social experience and still immature professional knowledge (20 students/doctors) and two groups with rich social experience and abundant knowledge (10 doctors). Specifically, participants were recruited from those currently receiving education at a university and from medical institutions affiliated with the university hospital. The group with lower social hierarchy consisted of 7 men and 13 women, with an average age of 23.85 years ($SD$ =4.38). The group with higher social hierarchy consisted of 7 men and 3 women, with an average age of 40.0 years ($SD$ =10.35). Each participant received 4,500 JPY on a QUO card. The study was conducted in a room equipped with video cameras and microphones.

The room was equipped with a table where participants could sit, a microphone array (Voice Compass produced by NTT-AT Co. Ltd.) used to drive the robot's actions that was placed at the center, a camera, and a stand for displaying discussion topics on posters. Participants were divided into groups of three, with one person from the higher hierarchy group(superior) and two from the lower hierarchy group(subordinate), to engage in problem-solving discussions. Each group conducted three discussions, with the order randomly assigned to one of the three study conditions: (1) hierarchical intervention condition, (2) general intervention condition, and (3) no movement Condition. These are independent variables.

In the hierarchical intervention condition, the robot was pre-programmed to understand the hierarchical structure among the participants, specifically identifying who the social superior and subordinates were. In this condition, the robot ignored the superior and performed follow actions and encouragement actions every minute for the other two participants. This was designed to promote the engagement of the most passive and socially lower-ranked members, thereby fostering equitable participation within the group. In the general intervention condition, the robot did not understand the social hierarchy in advance and performed follow-up and encouragement actions for all participants. This aimed to encourage the engagement of the most passive members of the group. In the no movement Condition, the robot was simply placed on the table and remained inactive throughout the study.

### 5.2 Group tasks

Each group was asked to complete three discussion tasks. Each task involved a topic related to the use of AI in medical settings, where participants had to propose a solution to potential issues arising from AI implementation in healthcare within 10 minutes. These topics were created on the basis of medical journals[20, 23] and news articles about case studies, which were cited accordingly. The order of the topics was determined randomly. The three topics were as follows: "How should the results of prognosis predictions made by machine learning be shared with patients?", "How should clinicians explain AI to patients who are reluctant to accept it?", and "How should AI intervene in the decision-making process for organ transplant allocations?"

Participants were given specific case studies related to these topics and were required to propose one solution to the problem. The topics were displayed on posters placed in a position visible to all three participants, who then discussed the topics on the basis of the provided information.

### 5.3 Procedure

All participants signed a consent form upon arrival. After agreeing to participate in the study, they completed a pre-study questionnaire. This included demographic data such as age, gender, and occupation, as well as their impressions of the other participants (e.g., whether they perceived them as superiors). They were then guided to the laboratory and asked to sit in one of the three seats arranged around the table. The layout of the laboratory is shown in Figure 4. After sitting at the table, participants were introduced to the robot as a "smart robot" and informed that the robot would be listening closely to the discussion and might move. After receiving an explanation about the method and duration of the discussion, the posters detailing the topics were displayed (the order of the topics was also determined randomly). Immediately after, the discussion began, and the robot was activated to respond to the participants' conversations according to the randomly assigned experimental condition (hierarchical intervention, general intervention, or no movement).

Participants were not given any explanation about how the robot would move. After the first discussion, they completed a questionnaire about that discussion. Then, the second topic was presented, and another 10-minute discussion was conducted, followed by another questionnaire. Finally, the same process was repeated for the third topic, after which participants provided responses in a free-response section and underwent debriefing. The order of the conditions was determined randomly for each group.

## 5.4 Measures

Our study measurements were designed to analyze team performance in group discussions and to understand conversational turn-taking.

*Speaking actions of superiors and subordinates.* The videos of each group's discussions were coded for task-related communication by participants at each hierarchical level. Non-task-related communication was discarded as it was minimal and participants were focused on the task. Task-related utterances were coded as speaking turns. Through this measurement, speaking turns and speaking time were calculated. A coding manual was created to code the videos, on the basis of the theories of Duncan [9]. According to these theories, only one person can control the conversation at a time, and this person is considered the "speaker." The speaker controls the direction of the conversation, while others who speak during this time are either backchanneling or competing for control of the conversation, in which case there is no clear speaker.

These coding schemes were used by a trained coder. Specifically, the speech analysis software Praat [3] was used to detect speaking intervals, and the speaker was assigned for each interval. When multiple people spoke simultaneously, the main speaker controlling the conversation was identified.

The speakers were then tagged according to their social hierarchy. The amount of speech and the number of turns were recorded for the person from the higher social hierarchy group, and for the two from the lower hierarchy group, the person with the longer speaking time and the person with the most turns were identified and their times recorded.

*Unevenness of speaking time and turns.* We created indices to represent the unevenness of speech in the group discussions (speaking time and turns). These were calculated for each group on the basis of the coded speech. These measurements reflected how skewed the speaking time and turn-taking were within a particular group. For each measurement, the group average (turns/time) was first calculated. Then, the difference between each individual member's turns/time and the group average was obtained. These differences were summed to reflect how balanced the group's speaking time/turn-taking was. The smaller the value, the more balanced the group, and the larger the value, the greater the imbalance. Specifically, the following formulas were used:

$$\text{turn} = \sum |t_i - \bar{t}|$$

where $t_i$ represents the number of turns taken in a single discussion, and $\bar{t}$ represents the average number of turns taken by the three members of the group.

$$\text{time} = \sum |p_i - \bar{p}|$$

where $p_i$ represents the number of seconds spoken in a single discussion, and $\bar{p}$ represents the average speaking time of the three members of the group.

*Questionnaire.* At the end of each discussion, participants were asked to rate their satisfaction with the discussion on a 6-point Likert scale, and their responses were collected and aggregated.

In this study, it was anticipated that superiors might be dominant and subordinates might find it difficult to speak up, leading to different qualities of satisfaction. Therefore, it was necessary to calculate the satisfaction of superiors and subordinates separately. The satisfaction of subordinates was averaged for the two subordinates, and the subordinate satisfaction for each group was calculated.

## 6 Results

First, as a manipulation check, we examined whether a hierarchical structure existed within the groups. In all groups, the participants recruited as superiors responded to the question "Do you think the person next to you is a superior?" with either "No" or "Not at all." Conversely, the subordinates responded to the same question with either "Yes" or "Very much so." This result indicates that a clear social hierarchy was observed in all groups. Next, we conducted a one-way ANOVA with three levels to examine whether there were differences in speaking time and number of turns between superiors and subordinates across conditions. Members could be divided into two hierarchical levels (superior and subordinate). As there were two subordinates, we took the average of the two for each group.

A one-way ANOVA with three levels using the superior's speaking time as the dependent variable showed a significant main effect of the robot's algorithm condition ($F(2, 18) = 4.49$, $p = .026$, partial $\eta^2 = .333$). Further post-hoc analysis using Shaffer's method indicated that the difference between the hierarchical intervention condition and the general intervention condition ($t(9) = 2.89$, adj.$p = .053$), and between the hierarchical intervention condition and the no movement condition ($t(9) = 1.90$, adj.$p = .089$) were marginally non-significant. There was no significant difference between the general intervention condition and the no movement condition ($t(9) = .643$, adj.$p = .536$). A one-way ANOVA with three levels using the subordinates' average speaking time as the dependent variable showed a marginally significant main effect of condition ($F = 3.3766$, $p = 0.0569$, partial $\eta^2 = 0.2728$). Post-hoc analysis using Shaffer's method showed no significant differences between the general intervention and hierarchical intervention conditions ($t = 2.3777$, adj.$p = 0.1241$), between the general intervention and no movement conditions ($t = 1.6986$, adj.$p = 0.1241$), or between the hierarchical intervention and no movement conditions ($t = 0.7445$, adj.$p = 0.4756$).

A one-way ANOVA with three levels using the superior's number of turns as the dependent variable showed no statistically significant main effect of condition ($F = 0.6193$, $p = 0.5494$, partial $\eta^2 = 0.0644$). Similarly, a one-way ANOVA with three levels using the subordinates' average number of turns as the dependent variable also showed no statistically significant main effect of condition ($F = 2.2526$, $p = 0.1339$, partial $\eta^2 = 0.2002$).

Furthermore, we conducted a one-way ANOVA with three levels to analyze the satisfaction levels of superiors across conditions,

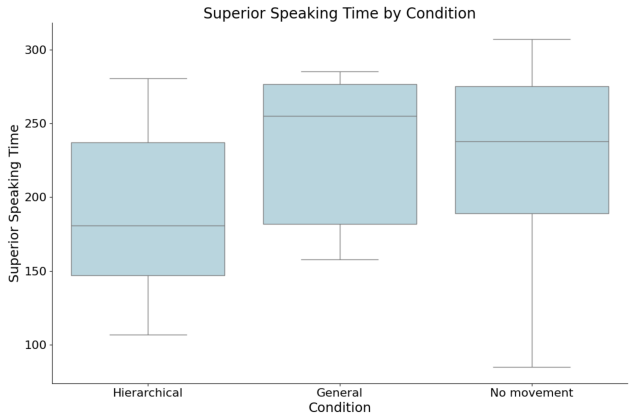

Figure 5: Experimental results: Regarding the speaking time of superiors. While a main effect of condition was observed, the differences in superiors' speaking time between the hierarchical condition and the other two conditions were marginally non-significant.

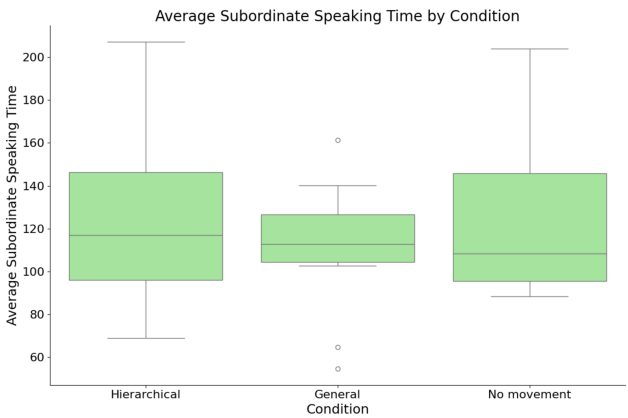

Figure 6: Experimental results: Regarding the speaking time of subordinates. The main effect of condition was marginally non-significant, and no significant differences were found in subordinates' speaking time between the hierarchical condition and the other two conditions.

but no significant effects were found ($F(2, 18) = 0.340$, $p = 0.716$, partial $\eta^2 = 0.036$). Additionally, a correlation analysis with the equality of speaking time and turns as explanatory variables and each member's discussion satisfaction as the dependent variable was conducted for each condition. The results showed that only in the No movement condition, the equality of speaking time significantly predicted the satisfaction of subordinates (subordinate1, subordinate2) ($r = .674$, $p = .032$).

## 7 Discussion
### 7.1 Hypothesis summary
In this experiment, subjective hierarchical relationships were observed in all groups, and under all conditions, it was found that the amount of conversation/number of turns of the superior was significantly higher. This indicates that the superior actively speaks and leads the conversation. Regarding H1, under the hierarchical intervention condition, we cannot conclude that the superior's conversation was reduced by taking action towards those considered to lack knowledge and experience in the social structure without intervening with the superior. While the main effect of the condition was significant, showing that the robot algorithm condition had an influence, post-hoc analysis between conditions narrowly failed to reveal significant differences between any conditions. Regarding H2, since the satisfaction of the superior did not change across conditions, it can be said to be generally supported.

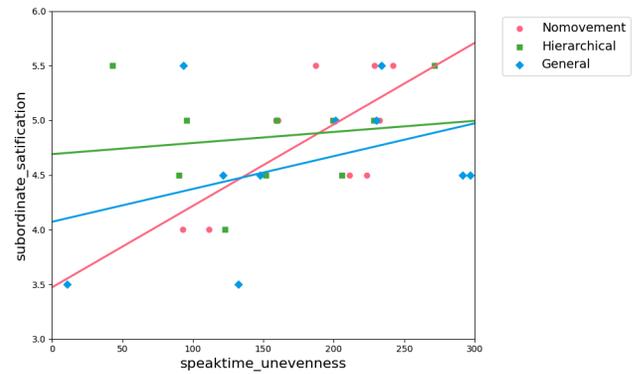

Figure 7: Experimental results: The correlation between the inequality of speaking time(Lower is better) and the satisfaction of subordinates. In the no movement condition, there is a significant correlation, while in the hierarchy and general conditions, no correlation is observed.

### 7.2 Effects of hierarchical intervention actions
In these groups with hierarchical structures, robots that understood the group's hierarchy and engaged in backchanneling behavior and encouragement of non-participating members demonstrated the potential to influence participation time among participants. This is evident from the observed main effect of conditions when examining the speaking time of superiors and subordinates. However, as there was no main effect of conditions on the number of turns for superiors and subordinates, no differences were observed between conditions in this respect. Focusing on the differences in speaking time between superiors and subordinates, the results of post-hoc tests showed that although no significant differences were found for either group, there was a tendency for superiors to have less speaking time in the hierarchical intervention condition compared to other conditions. When these effects are present, given that there was no main effect of conditions on the number of turns, it can be inferred that the decrease in speaking time was due to changes in speaking duration per turn rather than fewer opportunities to speak. It is not surprising that the length of their speech varied depending on conditions where the robot did or did not perform backchanneling or encouragement actions towards superiors. The way interventions are directed at superiors, such as whether or not backchanneling behavior is directed towards them, may cause them to feel unrecognized and alter their speaking time. Considering that the satisfaction of superiors did not significantly decrease under these conditions, this might be a method to suppress dominance without reducing their satisfaction. Drawing on these results, with more refined experiments, the development of an algorithm that can reliably suppress the speech of superiors might prevent situations where superiors dominate conversations and hinder information exchange. This could potentially be a method to address communication challenges in groups with hierarchical structures [12].

### 7.3 Effects of general intervention actions

In the current experiment, no changes in dynamics were observed when the robot intervened with all participants. This is because there were no significant differences in the equality of speaking time/turns within the group, or in the speaking time and turns of each individual (superior, subordinate), between the condition where the robot acted on all participants and the condition where the robot did not move. Previous studies have shown that the movements of vulnerable robots [27] or peripheral object robots [26] can increase participants' engagement in discussions in terms of speaking time and number of turns, but this difference was not observed in the current experiment. This may be because, in groups with a hierarchical structure, the ease of speaking is influenced by the hierarchy, and as a result, there was no observed effect on the increase or decrease in participants' engagement in the discussion.

### 7.4 Analysis of participant satisfaction

To further analyze the equality of speaking time, a test of no correlation was conducted for each condition to examine the correlation between individual satisfaction and the equality of speaking time. The results showed that the equality of speaking time was significantly correlated with the satisfaction of subordinates in the no movement condition, but no significant differences were observed in the other conditions. This indicates that in the no movement condition, where the robot did not intervene, the more equal the speaking time, the lower the satisfaction of the subordinates. This suggests that the robot's intervention may have alleviated the stress of speaking. One participant wrote the following in the free description section:

*Thanks to the robot prompting the conversation, it provided an opportunity to break the silence and made it easier to start talking.*

In conditions where the robot did not intervene, it took energy to start talking from silence or when prompted to speak, but with the robot present, it was possible to start conversations naturally, which may have influenced the satisfaction with the discussion.These findings are consistent with the results of robot intervention studies [13, 31], aimed at facilitating conversations in situations where it is difficult to start speaking.

### 7.5 Limitation and future work

In the robot design used in this study, only non-verbal behaviours and simple verbal actions were used. Whether a robot taking its own speaking turns and engaging in more complex dialogues can more directly influence the speaking turns within a group remains an unresolved issue. This is certainly an interesting area for future research. Practically, since real-time interactions are challenging, we believe it is crucial first to identify methods for group intervention using simple behaviour designs.

In the conditions where the robot intervened, the designed robot demonstrated two actions: "follow" and "encourage." These actions were set on the basis of pilot studies to match their respective purposes, but they were not selected from all possible action patterns. There are numerous variables, including the type and speed of actions and the angles of joint motors, which require validation across countless patterns. This also warrants further in-depth exploration.

Additionally, the SOTA robot was used. There is room to test other robots, and it is worth examining the impact of more realistic humanoid agents.

For this study, participants were recruited exclusively from a university hospital and a university. Therefore, the discussion topics were related to medical fields. This hierarchical structure of superiors and subordinates is specific to the medical department, and the results might differ in other settings such as companies or research labs. Particularly in the medical field, teams often exhibit hierarchical relationships [18], and it will be necessary to compare with groups that do not have such strong hierarchical structures in the future. Moreover, we did not systematically verify whether the participants knew each other, but since we recruited them individually, it is possible that participants had met previously in classes or on campus.

## 8 Conclusion

In the introduction of this paper, we emphasized that hierarchical structures are often observed in real-world problem-solving scenarios, and in such situations, subordinates may speak less and take fewer turns compared to superiors. Leveraging insights from the fields of HRI (Human-Robot Interaction) and CSCW (Computer-Supported Cooperative Work), we designed robot behaviors based on the social hierarchy within groups to address this issue. We proposed an approach to equalize participation in groups with social hierarchies by performing simple actions targeting subordinates while ignoring superiors, aiming to suppress the dominance of superiors, and conducted experiments to test this.

We found that actions considering the social hierarchy within groups and robot actions can influence the speaking time of superiors who tend to dominate conversations. However, to conclude whether actions considering social hierarchy can effectively suppress the dominance of superiors, which was our initial objective, more refined research is needed.

As robots become more prevalent, people will not only interact with robots one-on-one but also have more opportunities to interact with robots in group settings. Much of our daily lives are spent in small groups, and especially in problem-solving scenarios, there is potential for robot designs that improve discussions by considering social relationships within small groups. In this sense, it is essential to continue research that contributes to understanding how to consider social relationships and take appropriate actions. By doing so, we can deepen our understanding of what kinds of interactions promote better discussions in human-robot groups.

## 9  acks

This work was (partially) supported by JST, CREST (JPMJCR21D4), Japan.